\documentclass[sigconf]{acmart}

\makeatletter
\def\@mkbibcitation{}

\makeatother

\usepackage{amsmath} % For math environments
\usepackage{amsfonts} % For special math fonts like \mathbb
\makeatletter
\renewcommand{\thanks}[1]{\footnotetext[0]{#1}}
\let\@makefnmark\relax  % Removes the footnote mark
\makeatother

\usepackage{algorithm}
\usepackage[noend]{algorithmic}
\usepackage{xcolor}
\usepackage{booktabs} % For formal tables
\usepackage{graphicx} % Required for including images
\usepackage{array} 
\usepackage{algorithm}
\usepackage{subcaption}
\usepackage{graphicx}
\usepackage{textcomp}
\usepackage{xcolor}
\usepackage{multirow}
\usepackage{tabularx}
\usepackage{diagbox}
\usepackage{caption}

\usepackage{footnote} % For footnotes
\usepackage{fancyhdr}

\copyrightyear{2024}
\acmYear{2024}
\setcopyright{acmlicensed}\acmConference[DAC '24]{61st ACM/IEEE Design Automation Conference}{June 23--27, 2024}{San Francisco, CA, USA}
\acmBooktitle{61st ACM/IEEE Design Automation Conference (DAC '24), June 23--27, 2024, San Francisco, CA, USA}
\acmDOI{10.1145/3649329.3655665}
\acmISBN{979-8-4007-0601-1/24/06}

\acmPrice{}

\newcolumntype{Y}{>{\centering\arraybackslash}X}

\begin{document}

\title{Enabling On-Device Large Language Model Personalization with Self-Supervised Data Selection and Synthesis}

% Combined author line
\author{Ruiyang Qin$^{1}$, Jun Xia$^{1}$, Zhenge Jia$^{1}$, Meng Jiang$^{1}$, Ahmed Abbasi$^{1}$, Peipei Zhou$^{2}$, Jingtong Hu$^{2}$, Yiyu Shi$^{1}$}

\renewcommand{\shortauthors}{Qin et al.}

\affiliation{
    \institution{$^{1}$University of Notre Dame $^{2}$University of Pittsburgh}   
    \city{}
    \state{}
    \country{}
}

\begin{abstract}
After a large language model (LLM) is deployed on edge devices, it is desirable for these devices to learn from user-generated conversation data to generate user-specific and personalized responses in real-time. 
However, user-generated data usually contains sensitive and private information, and uploading such data to the cloud for annotation is not preferred if not prohibited. While it is possible to obtain annotation locally by directly asking users to provide preferred responses, such annotations have to be sparse to not affect user experience. 
In addition, the storage of edge devices is usually too limited to enable large-scale fine-tuning with full user-generated data. It remains an open question how to enable on-device LLM personalization, considering sparse annotation and limited on-device storage. In this paper, we propose a novel framework to select and store the most representative data online in a self-supervised way. Such data has a small memory footprint and allows infrequent requests of user annotations for further fine-tuning. To enhance fine-tuning quality, multiple semantically similar pairs of question texts and expected responses are generated using the LLM. Our experiments show that the proposed framework achieves the best user-specific content-generating capability (accuracy) and fine-tuning speed (performance) compared with vanilla baselines. To the best of our knowledge, this is the very first on-device LLM personalization framework.  
\end{abstract}

\maketitle
% \keywords{On-device learning, natural language processing, large language model (LLM), data selection}
\thanks{This work was supported in part by NSF 2324864, NSF 2122320, NSF 2324937, and NIH R01EB033387.}

\section{Introduction}
While most Large Language Models (LLMs) are still deployed in cloud servers, more and more LLMs (e.g. Llama-3B with parameter size of 6GB) are being deployed on edge and mobile devices such as prompt-driven robots to assist people in daily life \cite{brohan2023rt} and provide personalized companionship \cite{irfan2023between} while preserving users' privacy.
Traditionally, most LLMs are pretrained in high-performance servers and then deployed in these devices without further training. However, such a 
generic model usually falls behind in adapting to each individual user's 
unique needs and habits. It is often desirable for the deployed LLMs to further learn from real-world input data (e.g. user- and LLM-generated texts in their interaction), so that the LLMs can be personalized and adapted to the user's immediate context in real-time. This allows more accurate and context-aware responses, improving overall effectiveness of the LLMs.

% \thanks{This work was supported in part by NSF CNS-2007274.}

Although LLMs are pre-trained in a self-supervised way through next-token prediction, existing work has demonstrated that their fine-tuning must be supervised, where human-written outputs for task instructions or annotated outputs in a specific task dataset are given. For on-device 
personalization,  
it is usually impractical to send new data to the cloud for annotation due to data privacy and security concerns \cite{xu2023federated}. As such, any annotation has to be done locally by directly asking users to provide preferred responses during the user-LLM interaction. Such annotations need to be sparse because frequent inquires impede the user experience of LLM. Thus, for on-device LLM personalization, it is desirable to learn from new streaming data in-situ with as few annotations as possible.

In addition, for on-device personalization, considering limited hardware resources on the edge, it is necessary to learn from user-generated data streams without accumulating a large dataset. 
In other words, a small data buffer should be used to form each mini-batch for training. 
Existing LLM training methods assume that each mini-batch is independent and identically distributed (iid) by sampling uniformly at random from each semantic domain \cite{ding2023parameter}. 
However, it is challenging to maintain the most representative data in the buffer so that learning from this buffer efficiently derives a model that is as effective as if the entire data is used. 
This is due to two reasons. \textit{First}, the streaming data collected on edge devices are usually temporally correlated \cite{orhan2020self} and result in a correlation within each mini-batch. 
There can be a few rounds of uncontroversial dialogue sets before switching to those that contain useful information.  
\textit{Second}, there is no easy way to select representative data for each domain topic such that the data contain rich information in each domain topic from non-iid streaming data, due to the fact that the streaming data are {\em unlabeled}. 
If annotations were available for all the data,
we could easily select representative data 
based on all the annotations even if the streaming data were non-iid. 
Without addressing these challenges, directly learning from
temporally correlated non-iid mini-batches would result in poor representations and inefficient personalization.

To tackle the challenges of sparse local annotations and limited buffer size for on-device LLM personalization, in this paper, we propose to utilize embedding entropy, domain-related information, and embedding similarity to measure data quality from different perspectives in an unsupervised way. For each dialogue set in the data, the scores measured by the three metrics reflect the quality of the data regarding the information it contains as well as the domain it belongs to. Based on the three metrics, we propose a data replacement policy for the buffer, which always replaces the data in the buffer that has the lowest scores in these metrics if the buffer is full and the new data have higher scores. To provide annotation needed in the fine-tuning, we ask users to provide preferred responses as annotations for all the data in the buffer.    
Finally, multiple semantically similar question-answer pairs can lead to better model fine-tuning \cite{wei2021finetuned}. Therefore, for each dialogue set selected to store in the buffer, we utilize the LLM to synthesize semantically similar pairs, also without user supervision.

% Deep learning models have been increasingly deployed on personal devices due to multiple reasons, such as user-generated data privacy and model personalization. 

In summary, the main contributions of the paper include:
\begin{itemize}
    \item \textbf{On-device LLM personalization framework.} We propose a framework to form mini-batches of training data for fine-tuning LLM on the fly from the unlabeled input stream generated from user-LLM interactions. It only uses a small data buffer and eliminates the necessity of storing all the streaming data in the device.
    \item \textbf{Quality metrics for data selection.} We propose a data replacement policy guided by three quality metrics to maintain the most representative data in the buffer for on-device LLM fine-tuning. Annotation is not needed in the data replacement process.
    \item \textbf{Data synthesis for labeled pairs.} We propose to use the LLM model to generate additional data that are semantically similar to the selected data to enhance fine-tuning quality.
\end{itemize}

As this is the first work for on-device LLM personalization, no state-of-the-art is available, and we constructed a few vanilla baselines for comparison. Experimental results on multiple datasets of varying temporal correlation including ALPACA \cite{alpaca}, DOLLY \cite{DatabricksBlog2023DollyV2}, MedDialog \cite{chen2020meddiag}, Prosocial-Dialog \cite{kim2022prosocialdialog}, OPENORCA \cite{OpenOrca}, and Empathetic-Dialog \cite{rashkin-etal-2019-towards} show that the proposed framework achieves up to 38\% higher ROUGE-1 than the baselines and at the same time greatly improves the learning speed. %With 1\% annotated data on MedDialog dataset, the proposed framework achieves 1.39X rouge-1 improvement than direct apply LLM inference. Also with 1\% annotated data on MedDialog Dataset, the proposed framework improve rouge-1 up to 35\% compared to the SOTA data selection \cite{kirchhoff2014submodularity}. Meanwhile, the proposed framework achieves up to 3X faster learning speed than the baselines with higher rouge-1 obtained.
\begin{table}[h]
\centering
\smaller
% \footnotesize
\caption{Three example domains and their lexicons.}
\begin{tabularx}{\linewidth}{clY}
\toprule
\multicolumn{2}{l}{\textbf{Domain}} & \textbf{Example Lexicons} \\
\midrule
\multirow{3}{*}{\rotatebox[origin=c]{90}{\textbf{medical}}} & Admin & dose vial inhale inject  ml pills ingredient \\
                       & Anatomy & Pelvis arm sinus breast chest lymph tonsil\\
                       & Drug & ACOVA ACTONEL CARTIA EMGEL\\
\midrule
\multirow{3}{*}{\rotatebox[origin=c]{90}{\textbf{emotion}}} & Fear & bunker cartridge cautionary chasm cleave \\
                       & Surprise & amazingly hilarious lucky merriment \\
                       & Trust & advocate alliance canons cohesion \\
\midrule
\multirow{3}{*}{\rotatebox[origin=c]{90}{\textbf{GloVe}}} & GloVeTW26 & extreme potential activity impact movement \\
                       & GloVeCC41 & symptomatic thrombosis fibrillation \\
                       & GloVeTW75 & nyquil benadryl midol pepto midol ritalin\\
\bottomrule
\end{tabularx}
% \caption{Three Example domains and their lexicons}
\label{tab:example}
\end{table}

\section{Background and Related Work}

\subsection{Background}

\subsubsection{Text Domain}
Text domain usually refers to either the text topic like medical conversation or the embedding lexicon dictionary such as GloVe embedding dictionary. The lexicons related to certain text domains are organized as TABLE ~\ref{tab:example} shown. The medical, emotion and GloVe are three domains. In each domain, high-level lexicons such as \textit{fear} and \textit{drug} are used to index the detailed lexicons shown in \textit{Example Lexicons} in TABLE ~\ref{tab:example}.

\subsubsection{Text Embedding}
Text embedding involves converting text data into machine-readable numeric data. The quality of this embedding can influence subsequent text learning tasks and is also intrinsically linked to alignment—a crucial aspect of NLP \cite{dou2021word}. A prevalent embedding method assigns unique indices to words based on their position within a comprehensive vocabulary dictionary. Consequently, the same word, regardless of its occurrence in different text data, can be represented consistently using its unique index. In this work, we adopt an embedding technique using a pretrained transformer model. This model not only captures semantic information but also offers superior alignment capabilities.

\subsection{Related Work}

\subsubsection{LLM Personalization}

{LLM personalization} employs fine-tune the model to enhance its capability of text-understanding and text-generating in specific domains. While existing works concentrate more on scaling up the LLM to enable its comprehensive capabilities, some efforts \cite{ouyang2022training} have been made to fine-tune LLM using relative small dataset with high quality. However, all these works still involve large-scale computation and high-intensive neural network training with the overwhelming dataset size regarding on-device learning, and they assume that each mini-batch can be formed by sampling from the dataset. But when learning from streaming data, data is collected sequentially as it is. Random sampling from the entire input stream to create iid mini-batches is infeasible since it requires to store all the data, which is unavailable for device storage and computationally intractable for device computational resources. Therefore, an approach to form mini-batches on-the-fly while including the most representative data in each mini-batch is needed to enable efficient on-device LLM learning.

\subsubsection{Data Selection in Streaming and Continual Learning}

There are several supervised streaming and continual learning models that can learn from a stream of data \cite{aljundi2019gradient}. To overcome the problem of catastrophic forgetting of previously seen data, a data buffer is usually needed to store previous data for rehearsal \cite{wu2021enabling}. However these works cannot handle text input with user-specific semantic domains due to the lack of semantic level data processing and evaluation of these works. Efficiently evaluating input text data and selecting the most representative text which can shape the LLM towards user-specific text generation on devices have not been explored and studied.

\begin{figure}[h!]
  \centering
  \includegraphics[width=1\columnwidth]{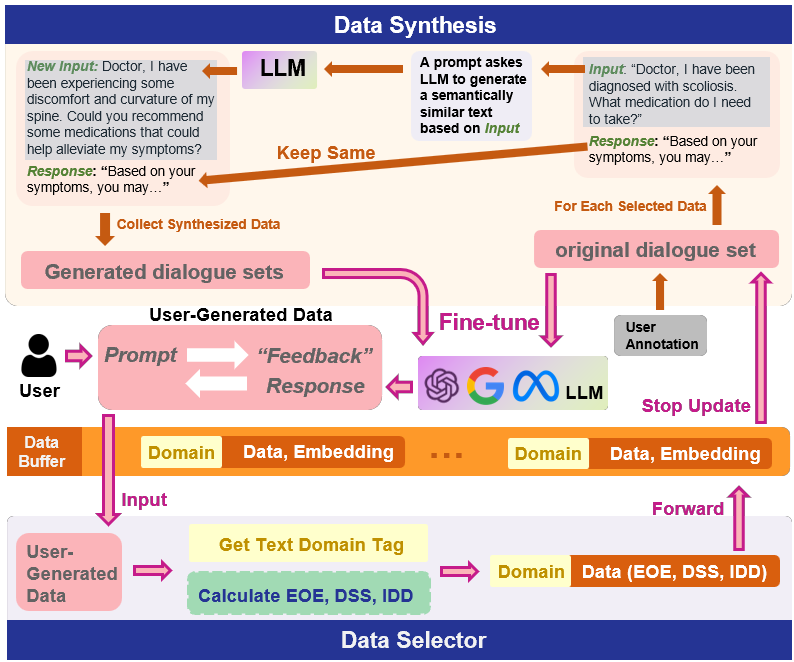}
  \caption{Overview of the framework. Fine-tune LLMs using data from data selection and following data generating.}
  \label{fig:overview}
\end{figure}
\section{Proposed work}

% This paper proposes a framework 
In this section, we first provide an overview of our framework. We then delve into the details, starting from the three metrics we have found to benefit most for the data selection in LLM personalization, and demonstrate how they collaborate with the data buffer to select data. After that, we demonstrate the data synthesis method we use to augment the selected data and explain the reason to use that.

\subsection{Framework Overview}
In our framework, we assume the atomic unit of data selection is a {\em dialogue set}, which contains a pair of question and answer during user-LLM interaction. 
%These metrics will be used to guide the data replacement policy for the on-device buffer. A self-supervised data synthesis method based on the selected data is also put in place to enhance the fine-tuning process. 

As shown in Figure~\ref{fig:overview}, the proposed framework has three stages. The first stage selects data to store in the data buffer based on certain quality metrics, and the selected data will be annotated by the user. The second stage takes the selected (and annotated) data and synthesizes additional data using the LLM. And finally, the selected and synthesized data together will be used for fine-tuning. In the discussion below, we focus on the first two stages where our framework resides. 

%At the core of our framework is an efficient and self-supervised way to evaluate each dialogue set in stream data, so that no personal data needs to be uploaded to the cloud and no frequent query of user annotation is needed. Towards this, we resort to information-theoretic metrics and pre-collect lexicons related to different domains such as medicine and emotional assessment, based on which we evaluate the quality of each dialogue set in the incoming stream for LLM fine-tuning. 

Specifically, with details discussed in Section 3.2, in the first stage, the proposed framework takes each dialogue set in the input streaming data from user-LLM interaction on-the-fly, calculate the quality metrics, and discard the data or update the data buffer based on the metrics. Considering the resource limitation, only a small data buffer is used to maintain the highest quality data. We will inquire user about the expected response as annotation for each selected dialogue set. 
%When the data buffer is full and a dialogue set of new input dialogue set $I$ arrives, both the new input data and the data in $B$ will be used to calculate the metrics of $I$. To compute the metrics, the device also maintains a pre-prepared dictionary formed with common lexicons in domains $L = \{l_1, l_2, \ldots, l_m\}$ where $m$ is the total number of possible domains that need to be covered. Examples of domains and lexicons are shown in TABLE~\ref{tab:example}. 

With details discussed in Section 3.3, in the second stage, each selected  dialogue set in the buffer is sent to the LLM for generation of additional dialogue sets that are semantically similar. We use the user annotation to replace the LLM generated response in the selected dialogue set. A pre-stored and fixed prompt is given to instruct the LLM for data generation. For the generated dialogue sets, a sanity check is made to make sure that their semantic similarity with the original dialogue set is above a user-specified threshold.   

\subsection{Data Selection by Quality Scores}

Each dialogue set's quality is captured by scores from three metrics. Each of them measures the quality of data from different perspectives as detailed below.

% \begin{figure}[t!]
%   \centering
%   \includegraphics[width=\columnwidth]{fig_mc_data_selector.png}
%   % \caption{Micro-Controller guides data selector and uses the three-step evaluator to update the data loader in data selector. }
%   \caption{Data Buffer. The EDI evaluator works with domain buffers to efficiently select data}
%   \label{fig:mc_data_selector}
% \end{figure}

% The lexicons from different domains are mutually exclusive and can be used to determine which domain the input data can be mostly related. The existing works using selective data to improve the on-device learning performance manage user input data via a single data buffer \cite{wu2021enabling}. In our framework, the inputs belong to the same domain can be repeatedly retrieved and evaluated. So we propose a designed data buffer structure to manage data efficiently.

% As shown in Figure ~\ref{fig:overview}, a data buffer is partitioned into multiple sub-buffers. Every sub-buffer has the same length. In each sub-buffer, input belongs to the same domain will be hold. Other than that, each sub-buffer is splitted into three sectors: primary sector, secondary sector, and cache sector. Such sectors are collaborated with the three metrics to accumulate high quality data:

\textbf{Metric 1: Entropy of Embedding.} Entropy of embedding (EOE) comes from the idea of Shannon's Entropy \cite{kapur1994measures}, where higher entropy means more features to learn. For each input $T$, EOE aims to qualify and measure the information of embedding vector $\vec{\mathbf{E}} = f(T)$ generated by an end-to-end embedding function $f(\cdot)$. The embedding vector $\vec{\mathbf{E}} = [e_1, e_2, \ldots, e_q]$ where $e_i$ is the embedding of the $i^{th}$ token in the input and $q$ is the length of the embedding. $EOE(\cdot)$ can then be defined as:

% Given an input, we want to first measure the amount of information it can contain. The input containing higher information can contain more features \cite{collobert2011natural} to be extracted and learned. Such information can be interpreted and measure using Shannon's Entropy \cite{kapur1994measures} from information theory. 
% % Essentially, entropy can provide useful descriptions \cite{yu2021information} of the underlying behavior of random variable
% Given the embedding $\vec{\mathbf{E}} = [e_1, e_2, \ldots, e_n]$ of input sequence containing tokens of $n$, the \textbf{E}ntropy \textbf{O}f \textbf{E}mbedding (\textbf{EOE}) can be calculated as:
% \begin{equation}
%     \text{EOE}(\vec{\mathbf{E}}_i) = \frac{-\sum_{e_i \in \vec{\mathbf{E}}} p(e_i) \log p(e_i)}{\log(n)}
% \end{equation}

\begin{equation}
    \text{EOE}(\vec{\mathbf{E}}_i) = \frac{-\sum_{e_i \in \vec{\mathbf{E}}} p(e_i) \log p(e_i)}{\log(n)}
\end{equation}

where \( p(e_i) \) represents the probability distribution of $e_i$, and $n$ is the number of tokens in $T$. 
The term \( p(e_i) \log p(e_i) \) represents the contribution of each token's embedding to the overall entropy. 
The normalization by \( \log(n) \) adjusts for the effect of the sequence length, ensuring that entropy is comparable across sequences of different lengths.

\textbf{Metric 2: Domain Specific Score.} While EOE measures the amount of information the input data contains, it cannot provide assessment regarding how much the information can be related to certain domains. As shown in TABLE~\ref{tab:example}, the domain of medical, emotion, and GloVe embedding can include distinct lexicons. The value of a dialogue set with respect to a particular domain can then be indicated by 
the token overlapping between the dialogue set and the common lexicons in each domain.  Note that this would require a pre-stored dictionary containing common lexicons of domains of interests in the device, which can be easily constructed. 
% and such measurement can be used in domain-specific text analysis \cite{bevilacqua2023automated}. 
Given a dialogue set $T$ containing $n$ tokens, and a collection of lexicon set $L = \{l_1, l_2, \ldots, l_m\}$ from $m$ different domains, the Domain Specific Score (DSS) can be calculated as:
\begin{equation}
    \text{DSS}(T, L) = \frac{1}{m} \sum_{i=1}^{m} \frac{|T \cap l_i|}{n}
\end{equation}
where it measures the ratio of tokens in $T$ belonging to every domain lexicons and output the mean of all ratios across all the domains. 
%\newcolumntype{L}{>{\raggedright\arraybackslash}X}
As domain can be highly important when adapting LLM to different tasks, the texts in different domains should not be compared to each other purely using EOE, and the text in the same domain should be evaluated together.

\textbf{Metric 3: In-Domain Dissimilarity.} While DSS calculates the general overlapping between $T$ and all domain lexicons, it is important to evaluate how much value $T$ brings to the domain it overlaps most with, 
i.e., the dominant domain. The dominant domain can be obtained as:
\begin{equation}
    \label{eq: dominant_domain}
    Dom_d = \underset{l_i \in L}{\arg\max} \, |T \cap l_i|
\end{equation}
When a dialogue set is stored in the buffer, we also store its dominant domain and its embedding. When a new dialogue set is considered, we identify all the 
dialogue sets already in the buffer that have the same dominant domain 
as the new set, and compute the dissimilarity between them, which will reflect the amount of new information that can be brought by the 
new set to the dominant domain. 
Specifically, the In-Domain Dissimilarity (IDD) can be calculated by cosine embedding similarity:
% \begin{equation}
%     IDD(E_i, S) = 
% \end{equation}
% With the domain determined, as shown in Figure ~\ref{fig:overview}, the input, along with its values of EOE and DSS, can be saved in cache sector of the sub-buffer to the corresponding domain. However, EOE and DSS measure the overall value of each input without considering the domain-wise differences of each selected input. 
% Due to the limited storage and computational resources, the data should contain high information with dissimilar meaning. Together, they can maintain a compact and high quality dataset. To measure the dissimilarity, we apply the cosine embedding similarity \cite{zhou2022problems} to compute the dissimilarity between inputs maintained within the sub-buffer.
\begin{equation}
    \label{eq: idd}
    \text{IDD}(\vec{\mathbf{E}}, B) = \frac{1}{R} \sum_{i=1}^{R} (1 - \cos(\vec{\mathbf{E}}, \vec{\mathbf{E}}^i_{Dom_d}))
\end{equation}
where $\mathbf{E}^i_{Dom_d}$ is the embedding vector of the $i^{th}$ dialogue set in the buffer $B$ that has the same dominant domain as $T$, and $R$ is the total number of such dialogue sets in $B$. \(\cos(\vec{\mathbf{E}}, \vec{\mathbf{E}}^i_{Dom_d})\) is the cosine similarity between \(\vec{\mathbf{E}}\) and \(\vec{\mathbf{E}}^i_{Dom_d}\), calculated as:
\begin{equation}
    \cos(\vec{\mathbf{E}}, \vec{\mathbf{E}}^i_{Dom_d}) = \frac{\vec{\mathbf{E}} \cdot \vec{\mathbf{E}}^i_{Dom_d}}{\|\vec{\mathbf{E}}\| \|\vec{\mathbf{E}}^i_{Dom_d}\|}
\end{equation}
Note that we store the embedding of all the selected dialogue sets 
in the buffer, so that they do not need to be re-computed each 
time a new dialogue set is being evaluated. 

% Given the embedding \(\vec{\mathbf{E}}\), its dissimilarity compared to a set of embeddings \(S = \{\vec{\mathbf{E}}_1, \vec{\mathbf{E}}_2, \ldots, \vec{\mathbf{E}}_j\}\), namely the \textbf{I}n-\textbf{D}omain \textbf{D}issimilarity (\textbf{IDD}), can be calculated as:
% \begin{equation}
%     \label{eq: idd}
%     \text{IDD}(\vec{\mathbf{E}}, S) = \frac{1}{j} \sum_{i=1}^{j} (1 - \frac{\vec{\mathbf{E}} \cdot \vec{\mathbf{E}}_i}{\|\vec{\mathbf{E}}\| \|\vec{\mathbf{E}}_i\|})
% \end{equation}

\textbf{Quality Score Based Data Selection.} When a new dialogue set arrives and the buffer is full, we need to decide whether this new set needs to be discarded, or to replace a dialogue set already in the buffer. If the latter, we also need to decide which set in the buffer needs to be replaced. In our framework, for each new input dialogue set $T$, its EOE, DSS, and IDD scores will be computed and compared with these scores of all the data in the buffer. If all the three metrics of $T$ are higher than a dialogue set already in the buffer, then we use $T$ to replace it. Note that if there are more than one options to replace, we will randomly select one. Users will then be asked to provide annotation to this new dialogue set, for example, by asking ``{\em Do you think my response is acceptable and if not what would be an ideal response?}'' If users provided an alternative response that is preferred, the dialog set will be updated
using the user provided content before being placed into the buffer. 

Finally, from the definition of the three metrics and the replacement policy, it is easy to see that for each new dialogue set, 
our data selection policy has a linear complexity with respect to the size of the buffer. 

\subsection{Data Synthesis}
The selected data in the buffer can capture features unique to the user. However, when such data are used in LLM fine-tuning, the limited size can confine the effectiveness. To address this problem, inspired by the observation that multiple semantically similar question-answer pairs can lead to better model fine-tuning\cite{wei2021finetuned}, we deploy a self-generated instruction strategy to generate additional data. 

Specifically, each dialogue set (i.e., ``original'' dialogue set) in the buffer will be sent to the LLM to generate similar ones, by giving the following prompt ``{\em Please refine and generate a text semantically similar
to the following text block, no need to answer it, no
need to explain, use [ ] to hold your generated response: }'' 
followed by the 
original dialogue set. We run this multiple times to generate several additional sets for each original one. To avoid complicating the data replacement, the data synthesis process will only occur right before the fine-tuning starts each time. 
%An example of an original dialogue set and two generated dialogue sets 
%are shown in Figure~\ref{XX}. 

However, sometimes we find that the dialogue sets generated by LLM, even 
though the prompt instructs it to generate semantically similar ones, 
still differ from the original dialogue set significantly, if measured by 
{\em ROUGE-1}. As such, we add a sanity check for each generated dialogue set, 
and if {\em ROUGE-1} between it and original set is above a threshold, it 
will be discarded.

\section{Experimental Evaluation}

\begin{figure*}[htbp]
  \centering
  % First row of figures
  \begin{subfigure}[b]{0.29\textwidth}
    \includegraphics[width=0.9\textwidth]{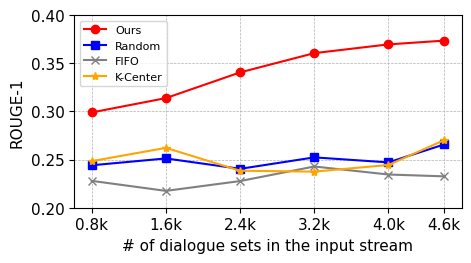}
    \caption{ALPACA}
    \label{fig:lc_1}
  \end{subfigure}
  % \hfill % space between the subfigures
  \begin{subfigure}[b]{0.29\textwidth}
    \includegraphics[width=0.9\textwidth]{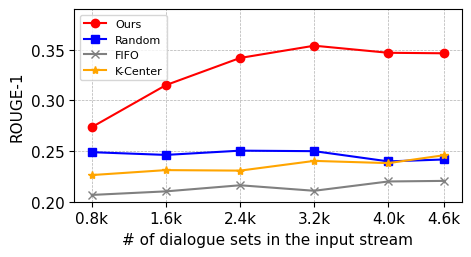}
    \caption{DOLLY}
    \label{fig:lc_2}
  \end{subfigure}
  % \hfill % space between the subfigures
  \begin{subfigure}[b]{0.29\textwidth}
    \includegraphics[width=0.9\textwidth]{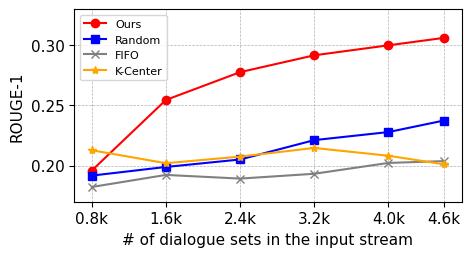}
    \caption{Prosocial-Dialog}
    \label{fig:lc_3}
  \end{subfigure}
  % Second row of figures
  \begin{subfigure}[b]{0.29\textwidth}
    \includegraphics[width=0.9\textwidth]{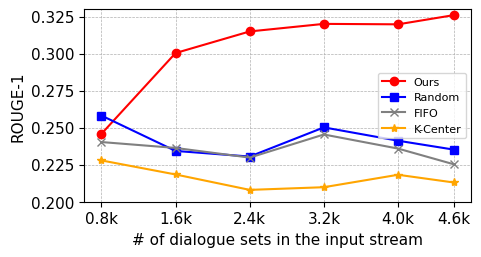}
    \caption{Empathetic-Dialog}
    \label{fig:lc_4}
  \end{subfigure}
  % \hfill % space between the subfigures
  \begin{subfigure}[b]{0.29\textwidth}
    \includegraphics[width=0.9\textwidth]{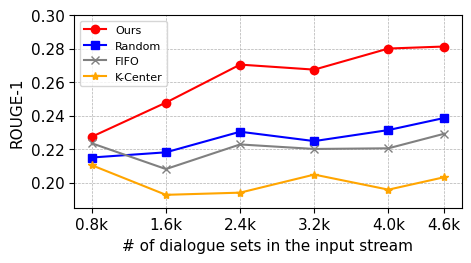}
    \caption{OPENORCA}
    \label{fig:lc_5}
  \end{subfigure}
  % \hfill % space between the subfigures
  \begin{subfigure}[b]{0.29\textwidth}
    \includegraphics[width=0.9\textwidth]{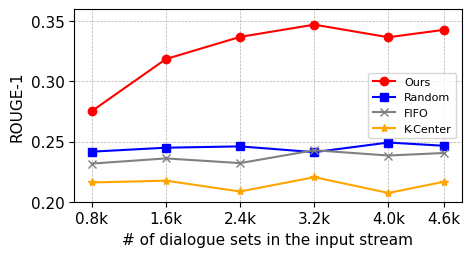}
    \caption{MedDialog}
    \label{fig:lc_6}
  \end{subfigure}
  \caption{The learning curve of our proposed framework, Random Replace, FIFO Replace, and K-Center with buffer size 281KB on datasets (a) ALPACA (b) DOLLY (c) Prosocial-Dialog (d) Empathetic-Dialog (e) OPENORCA (f) MedDialog.}
  \label{fig:lc_series}
\end{figure*}

\subsection{Experimental Setup}
We first explain the datasets used in the experiments, the settings under different experiments, and baselines.

\textbf{Datasets.} To show the generalization capability of our framework, we use multiple and diverse datasets, including ALPACA \cite{alpaca}, DOLLY \cite{DatabricksBlog2023DollyV2}, MedDialog \cite{chen2020meddiag}, Prosocial-Dialog \cite{kim2022prosocialdialog}, OPENORCA \cite{OpenOrca}, and Empathetic-Dialog \cite{rashkin-etal-2019-towards}, to evaluate the proposed framework. These datasets reflect different temporal correlation scenarios in the input data stream: ALPACA, DOLLY and OPENORCA contain diversified dialogue sets not bounded to a single domain, and the input data streams formed on them have little temporal correlation. While the other three ones are domain-specific, and thus the data streams are highly temporal correlated. All these datasets are fully annotated. However, our framework only uses annotations for the data selected to finetune the LLM; and the fully annotated dataset is used in the evaluation.

\textbf{Default Experimental Setting.} We use a pre-trained Llama-3B \cite{openlm2023openllama}, one of the most popular on-device LLM, as the model embedded on devices. For each dataset, we randomly choose 10\% of the data to simulate input data stream and run our 
framework on it for model 
fine-tuning, and the remaining 90\% is reserved for evaluation of the fine-tuned mode. 
For every 800 dialogue sets received in the input stream, we will start 
the fine-tuning process with 100 epochs using optimizer AdamW. The buffer will not be cleared after the fine-tuning, and the data selection continues after the fine-tuning is done. 
We obtain input text embedding from Llama-3B last hidden layer during its inference. Unless otherwise mentioned, in data synthesis each dialogue set in the buffer is sent to LLM to generate three additional sets. With the selected and synthesized data, we fine-tune Llama-3B using Low-Rank Adaptation (LoRA) \cite{hu2021lora}, a parameter efficient fine-tune technique. Unless otherwise specified, the batch size is 128 with fixed learning rate of 0.0003. For LoRA settings, the trainable layers are the QKV layers (\textit{q\_proj}, \textit{k\_proj}, \textit{v\_proj}) and attention output layer (\textit{o\_proj}), max sequence length is 512, LoRA rank \textit{r} is 8, loRA metrics scaling facotr \textit{alpha} is 16, and LoRA dropout rate is 0.05. For consistency, when we use the fine-tuned model to generate text for evaluation, the temperature $\tau$ is set to 0.5 for all experiments.  

As for the data selection buffer design, 
for efficient memory operations, we divide it
into bins of equal size and each bin is able to hold the text of one dialog set, its domain as well as its embedding. Considering that the maximum dialogue set is of length 1,024 tokens (512 tokens x2) and the embedding is a floating point vector of length 4,096 for Llama-32B, the bin size is set to 22KB. In the experiments, we will explore the impact of the buffer size. To make sure that our framework can be 
applied to the various edge devices, we explore buffer sizes from 
32 bins (704KB) to 512 bins (11MB). 
% We profiled the fine-tuning of the LLM model on the device of Nvidia Jetson AGX Orin. For one epoch fine-tuning, it can take 89.5s with 10W power consumption . 
To efficiently evaluate our framework, we use compact, 150 watt, single-slot A10 GPU, which is much smaller than 300 watt double-width A100 GPU. A10 is compatible to fit into robotics applications.
% We conducted our expeirments on both Nvidia Jetson AGX Orin and A10 GPU, which are compactable to fit into robotics applications. For AGX Orin, one epoch fine-tuning can take 89.5s with 10W power consumption.
% All the experiments are obtained on Nvidia Jetson AGX Orin.

\textbf{ROUGE-1 as Evaluation Metric.} After the LLM model is 
fine-tuned using our framework, for each dialogue set in the 
test set, we feed the same user question to the model and collect the 
response generated. The quality of the data can then be evaluated by measuring the overlapping between the generated responses and the responses in the test dialogue set under the same question, which can be captured by ROUGE-1. ROUGE-1 is commonly used in natural language processing to measure the overlap of unigrams (single words) between the machine-generated summary and a reference summary in terms of F-1 score \cite{eyal2019question}. A higher ROUGE-1 score suggests that the generated text is more similar to the reference text. 

\begin{table}[h]
\caption{ROUGE-1 of different methods on six datasets with data buffer 2816KB}
\centering
\footnotesize	
% \smaller
\begin{tabularx}{\columnwidth}{cYYYY}
\toprule
            & Random & FIFO   & K-Center & Ours \\ 
\midrule
ALPACA      & 0.2457 & 0.2013 & 0.2384 &   0.3736 \\
DOLLY       & 0.2417 & 0.1976 & 0.2403 &  0.3465 \\
Prosocial   & 0.2375 & 0.2190 & 0.2147 &  0.3062 \\
Empathetic  & 0.2352 & 0.1902 & 0.2098 &  0.3260 \\
OPENORCA    & 0.2286 & 0.1833 & 0.2048 &  0.2813\\
MedDialog   & 0.2465 & 0.2074 & 0.2204 &  0.3429 \\
\bottomrule
\end{tabularx}
\label{tab:baselines}
\end{table}

\textbf{Baselines.} As this is the first work on on-device LLM personalization, we do not have state-of-the-art for comparison. As such, 
we construct a few vanilla baselines. \textit{Random Replace} is recently used for continual learning \cite{hayes2019memory}. It selects data uniformly at random from new data to replace the ones already in the buffer. \textit{FIFO Replace} is also recently employed for continual learning \cite{hayes2019memory}. It replaces the oldest data in the buffer with new data. \textit{K-Center} is a SOTA active learning approach \cite{sener2017active} which selects the most representative data by performing
k-center clustering in the features space. While not directly used in LLM personalization, these works also do not require labeling
information and seemingly simple, and have
demonstrated superior performance in maintaining image data for
continual learning. In addition, to demonstrate the 
importance to consider all the three metrics EOE, DSS and IDD, we will perform ablation study on additional three baselines, each only using one of the three for data selection. 
For fair comparison, 
for all of these methods we used the same data synthesis based on the selected data as used in our framework. 

\subsection{Results}
We start with comparing the ROUGE-1 of Random Replace (Random), FIFO Replace (FIFO), and K-Center on 
all the datasets using buffer size 128 bins (2816 KB). The results are presented in TABLE~\ref{tab:baselines}. 
From the table we can see that our method outperforms all the baselines
by a significant margin, indicating that its superiority in both 
weak and strong temporal correlation settings. The results also show that the most competitive baseline is the
seemingly simple, yet surprisingly effective approach
random replace. These results match the
results in \cite{borsos2020coresets} for image classification tasks, where a random replacement policy outperforms
elaborately designed approaches.

\begin{table}[h]
% \smaller
\caption{ROUGE-1 based on MedDialog with different buffer sizes.}
\centering
\footnotesize	
% \smaller
\begin{tabularx}{\columnwidth}{ccYYY}
\toprule
Buffer Size (KB)& Ours              & Random    & FIFO      & K-Center     \\ \midrule
176             & 0.3040            &0.2281     & 0.2383    & 0.2160\\
352             & 0.3447            & 0.2455    & 0.2304    & 0.2175      \\
704             & 0.3353            & 0.2536    & 0.2389    & 0.2080    \\
1408            & 0.3353            & 0.2791    & 0.2417    & 0.2204     \\
2816            & 0.3940            & 0.2638    & 0.2309    & 0.2073     \\
5632            & 0.3944            & 0.2748    & 0.2381    & 0.2167      \\
11264           & 0.4215            & 0.2834    & 0.2315    & 0.2122      \\
\bottomrule
\end{tabularx}
\label{tab:buffer}
\end{table}

Next, as a very important profiling tool for on-device learning, we evaluate the learning curve of the proposed framework and the baselines on these datasets. The learning curve represents how well the LLM can be fine-tuned to generate user-specific text with respect to the number of input dialogue sets seen as the data streams in. The same buffer size is used. The results are depicted in Figure ~\ref{fig:lc_series} (a)-(f), respectively. From all the figures, we can clearly see that the ROUGE-1 of the proposed framework consistently increases with the increase of seen data, while the ROUGE-1 of the baselines only demonstrate minor improvement.

In addition, we evaluate the impact of buffer size on the performance of the proposed framework. The model is trained on the MedDialog dataset. The number of bins in the buffer is in \{8, 16, 32, 64, 128, 256, 512\} corresponding to a buffer size of \{176KB, 353KB, 704KB, 1408KB, 2816KB, 5632KB, 11264KB\} respectively. The corresponding learning rate is scaled to \{2, 3, 4, 5, 7, 10, 14\} X $10^{-5}$, roughly following a learning rate $\propto \sqrt{\text{batch size}}$ scaling scheme. The proposed framework consistently outperforms the baselines under different buffer sizes. As shown in TABLE~\ref{tab:buffer}, under the different buffer sizes, the ROUGE-1 by the proposed framework maintains a clear margin over the baselines. Besides, the margin becomes larger as the buffer size increases. This is because a larger buffer size provides the framework a better opportunity to select more high quality data, and the framework can leverage this opportunity to maintain richer quality data for learning, while the baselines cannot. Also, the proposed framework achieves higher ROUGE-1 when the buffer size becomes larger. This is because a larger buffer size provides a larger batch size, which naturally benefits the LLM fine-tuning.

% \begin{table}[h]
% \caption{ROUGE-1 of our framework and the baselines using only one of the three metrics EOE, DSS or IDD on six datasets with buffer size 2816KB.}
% \centering
% \footnotesize	
% \begin{tabular}{ccccc}  % Five columns now
% \toprule
%             & EOE & DSS & IDD & Ours \\ 
% \midrule
% ALPACA      & 0.2821 & 0.2726 & 0.2950 & 0.3736 \\
% DOLLY       & 0.2782 & 0.2633 & 0.2247 & 0.3465 \\
% Prosocial   & 0.2617 & 0.2441 & 0.2324 & 0.3062 \\
% Empathetic  & 0.2661 & 0.2726 & 0.2707 & 0.3260 \\
% OPENORCA    & 0.2468 & 0.2362 & 0.2468 & 0.2813 \\
% MedDialog   & 0.2608 & 0.2726 & 0.2931 & 0.3429 \\
% \bottomrule
% \end{tabular}
% \label{tab:ablation}
% \end{table}

% \DeclareCaptionLabelFormat{customlabel}{Table #2:}
% \captionsetup[table]{
%   labelsep=space,
%   labelfont={bf,small},
%   textfont=small,
%   labelformat=customlabel  % Use the custom label format
% }
\begin{table}[h]
    \caption{ROUGE-1 of our framework and the baselines using only one of the three metrics EOE, DSS or IDD on six datasets with buffer size 2816KB.}
    \label{tab:ablation}
    \centering
    % \smaller
    \footnotesize	
    \begin{tabularx}{\columnwidth}{cYYYY}
        \toprule
                    &EOE & DSS & IDD & Ours \\ 
        \midrule
        ALPACA      & 0.2821 & 0.2726 & 0.2950 &   0.3736 \\
        DOLLY       & 0.2782 & 0.2633 & 0.2247 &  0.3465 \\
        Prosocial   & 0.2617 & 0.2441 & 0.2324 &  0.3062 \\
        Empathetic  & 0.2661 & 0.2726 & 0.2707 &  0.3260 \\
        OPENORCA    & 0.2468 & 0.2362 & 0.2468 &  0.2813\\
        MedDialog   & 0.2608 & 0.2726 & 0.2931 &  0.3429 \\
        \bottomrule
    \end{tabularx}
\end{table}

Finally, we perform two ablation studies. The first one is to demonstrate the advantage of simultaneously considering 
all the three quality metrics EOE, DSS and IDD, 
we modify our framework to use only one of them for data replacement,  
those only considering one of them. The results on all six datasets 
are presented in TABLE ~\ref{tab:ablation}. From the table we can see that simultaneously 
considering all the metrics always achieves the highest ROUGE-1. 

\begin{figure}[h]
  \centering
  \includegraphics[width=0.89\columnwidth]{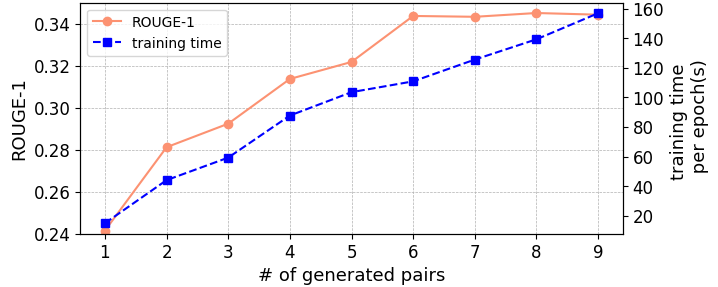}
  \caption{ROUGE-1/training time on MedDialog dataset with different number of dialogue sets generated from each original set in the buffer.}
  \label{fig:syn}
\end{figure}

The second study shows the relationship between the number of additional sets generated during data synthesis for each original dialogue set in the buffer and ROUGE-1/training time per epoch.
% The second study shows the relationship between ROUGE-1 and the number of additional sets generated during data synthesis for each original dialogue set in the buffer. 
From Figure ~\ref{fig:syn} we can see that the maximum gain in ROUGE-1 can be attained when six additional sets are generated, while the training time consistently increases. Generating more then six dialogue sets will not further boost the performance, but would cost more time to fine-tune the model.
% storage to hold the generated data.
For the sake of balanced efficiency and preference , as mentioned in the experimental setup, in all the experiments we generated three additional dialogue sets.
\section{Conclusion}
In this paper, we present a novel framework for on-device personalization of a large language model (LLM) on edge devices. Our approach addresses privacy concerns by selecting and storing representative data locally in a self-supervised manner. In addition, it uses semantically similar pairs of question texts and expected responses generated by LLM to enhance on-device learning performance. Our framework minimizes the need for frequent user annotations, and overcomes the challenge of sparse on-device storage. Experimental results show that our framework achieves superior user-specific content generation accuracy and fine-tuning speed compared to vanilla baselines. This paper marks the first on-device LLM personalization framework.

% \begin{acks}
% \end{acks}

\bibliographystyle{ACM-Reference-Format}
\bibliography{citations}

\end{document}